\documentclass[10pt,twocolumn]{article} 
\usepackage{latex8}
\usepackage{times}
\usepackage{graphicx}
\usepackage{color}
\usepackage{booktabs}
\usepackage{color}
\usepackage{multirow}
\usepackage{amsmath}
\usepackage{graphicx}
\usepackage{amsfonts}
\DeclareMathOperator*{\argmax}{arg\,max}

\usepackage[
  separate-uncertainty = true,
  multi-part-units = repeat
]{siunitx}

\begin{document}

\title{Data-specific Adaptive Threshold for Face Recognition and Authentication}
       
\author{Hsin-Rung Chou, Jia-Hong Lee, Yi-Ming Chan, and Chu-Song Chen\\
Institute of Information Science, Academia Sinica, Taipei, Taiwan\\
MOST Joint Research Center for AI Technology and All Vista Healthcare\\
$\{sherry18030, honghenry.lee, yiming, song\}@iis.sinica.edu.tw$
}

\maketitle
\thispagestyle{empty}

\begin{abstract}
Many face recognition systems boost the performance using deep learning models, but only a few researches go into the mechanisms for dealing with online registration. Although we can obtain discriminative facial features through the state-of-the-art deep model training, how to decide the best threshold for practical use remains a challenge. We develop a technique of adaptive threshold mechanism to improve the recognition accuracy. We also design a face recognition system along with the registering procedure to handle online registration. Furthermore, we introduce a new evaluation protocol to better evaluate the performance of an algorithm for real-world scenarios. Under our proposed protocol, our method can achieve a 22\% accuracy improvement on the LFW dataset.
\end{abstract}

\Section{Introduction}
Deep Convolutional Neural Networks~(CNNs) have achieved great success for face recognition. Especially after the unified framework proposed by Schroff \textit{et al.}~\cite{schroff2015facenet}.
In most works~\cite{chen2018mobilefacenets,parkhi2015deep,schroff2015facenet}, researchers use a fixed threshold for the face verification tasks. The threshold is usually the optimal value that can separate different identities of the testing data. However, we argue that this method is deficient for the real-world scenarios, because the optimal threshold is usually case-specific, \textit{i.e.} the best thresholds for different data sets are often different. As we do not have the testing data in practical applications, the optimal threshold is hard to find or even unobtainable. Furthermore, The content of a face database would change frequently, which also raise the need for threshold value tuning.

We propose the technique of feature-specific adaptive thresholding to improve the recognition accuracy. The adaptive threshold serves for two purposes: it performs the task of face verification, and it acts as a gatekeeper of the database. We also design a system to simulate the real-world scenario, which consists of a deep CNN and a database; we use the deep CNN to extract the embedded feature vector of a facial image and store it in the database along with the threshold and its identity.

To have a fair comparison with the results of using a fixed threshold, we introduce a new evaluation protocol that has the same registration flow as the proposed system. The experimental results show that our method outperforms the traditional one. It strengthens the robustness of the threshold and makes the selection process more tractable.

\Section{Related Work}
Recently, deep CNNs play an important role in the research of face recognition. These works attempt to obtain a better deep learning model by either modifying the objective functions or redesigning the network structures. Their general goal is to utilize the deep learning model to produce discriminative facial features.
For example, FaceNet~\cite{schroff2015facenet} proposes triplet loss; NormFace~\cite{wang2017normface} optimizes cosine similarity directly; A-Softmax loss~\cite{liu2017sphereface} introduces the angular margin into the objective and AM-Softmax loss~\cite{wang2018additive} improves A-Softmax to stabilize the training; VGGFace\cite{parkhi2015deep} combines VGG-net~\cite{simonyan2014very} with triplet loss and achieves a great success on the LFW benchmark; Recently, MobileFaceNets~\cite{chen2018mobilefacenets} designs a lightweight face recognition model with Global Depthwise Convolution.

Nevertheless, it is still challenging to choose the optimal threshold for differentiating different identities. In the benchmark of LFW~\cite{huang2008labeled}, a canonical list of face verification pairs is provided and researchers are asked to evaluate the algorithm via 10-fold cross validation (CV). The best threshold for $L_2$-distance is then decided accordingly. 
The CV-based methods are sufficient for comparing the experimental results of different methods, but the thresholds selected usually do not suffice for practical applications.
In this paper, instead of applying a fixed threshold to the entire database, we introduce an adaptive thresholding method that assigns a suitable threshold per each registered face in the database, and show that the approach performs more favorably on face recognition and authentication.

\Section{Methodology}
We have two operations in our system: registration and recognition. In the operation of registration, a feature vector (or embedding) is extracted form an input face image by using a deep network. We assume that one face image is registered on the system at a time. The face could belong either to some person already registered on the system, or a new person not registered before. 
We assign a threshold to the registered face during each registration, and the thresholds of the other registered faces will be modified accordingly.

\begin{figure}[t]
\centering
\includegraphics[scale=0.2,keepaspectratio]{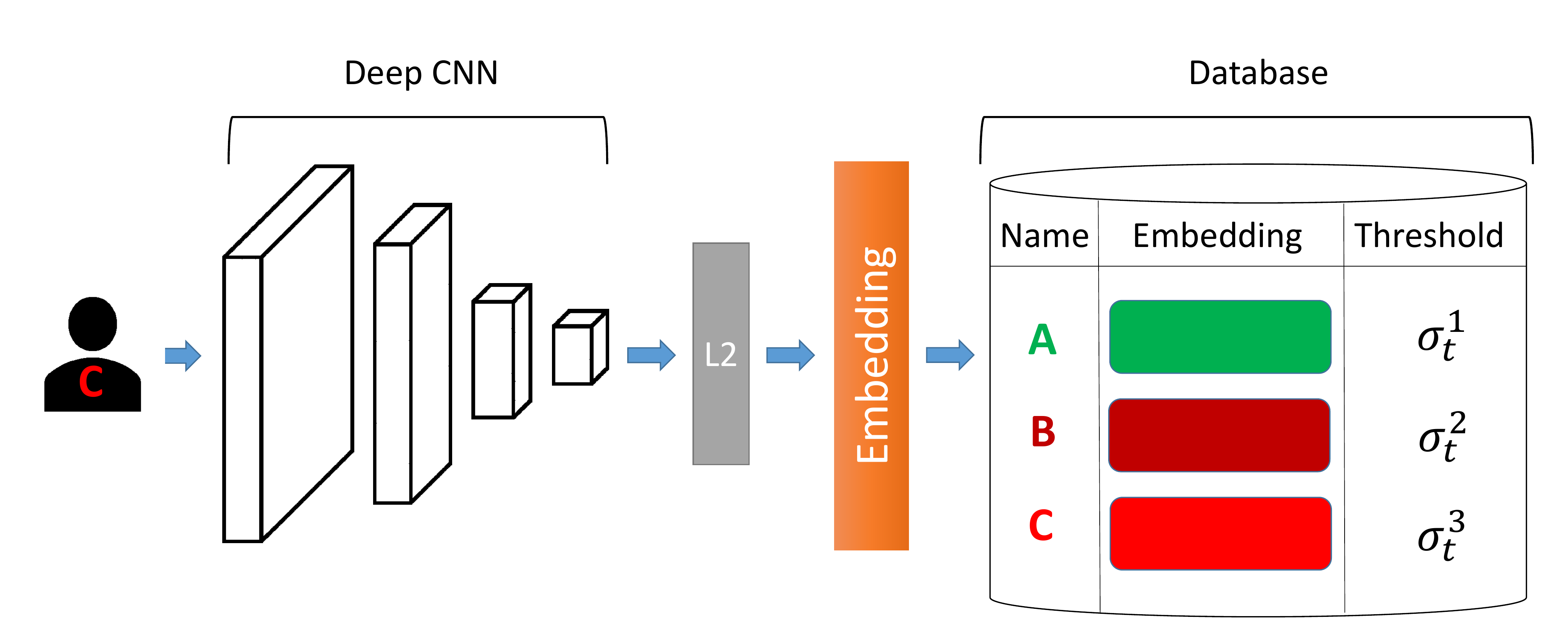}
\caption{System Structure. It consists of a deep CNN with a $L_2$ normalization layer, and a database for storing feature embeddings.}
\label{fig:structure}
\end{figure}

For recognition, given a query image, we extract its feature embedding and compute the similarity scores between it and all of the other stored embeddings. Then we use the similarity scores to determine the identity of the query image.
The system structure is illustrated in Figure~\ref{fig:structure}.

In the following sections, we first review the procedure of extracting the facial features by using state-of-the-art face detection and recognition methods; then we describe the details of registering and recognizing an image with adaptive threshold in sections 3.2 and 3.3.

\SubSection{Deep Convolutional Neural Network}

Given a query image, we first utilize the Multi-task Cascaded Convolutional Networks (MTCNN)\cite{zhang2016joint} to detect and align the face.
Next, we utilize a well-trained face recognition model trained on Inception-ResNet-v1\cite{szegedy2017inception} with a $L_2$ normalization layer as proposed by FaceNet\cite{schroff2015facenet}. In the inference phase, we extract the output of the $L_2$ normalization layer as the unified facial feature embedding.
For the following section, we use \textit{embedding} as a shorthand for the extracted facial features as introduced by FaceNet.
As the embeddings are $L_2-$normalized, we use the inner product between two embeddings to compute their similarity.

\begin{figure}[t]
\centering
\includegraphics[scale=0.2,keepaspectratio]{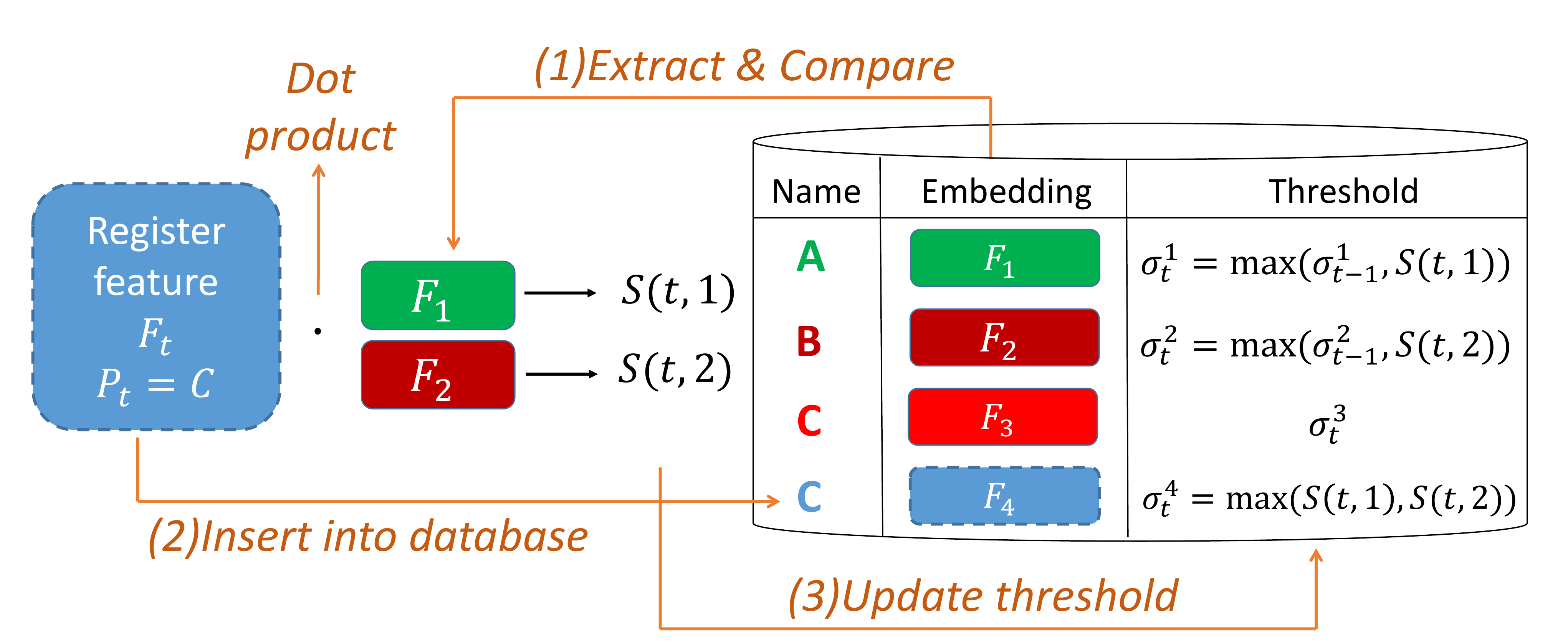}
\caption{Registration Flowchart. $F_{t}$ is the embedding of a registering image with identity $C$. (1) Compute the similarity scores of $F_{t}$ and all embeddings other than $C$. (2) Store $F_{t}$ and its name in the database. (3) Update the thresholds of all the embeddings accordingly.}
\label{fig:register}
\end{figure}

\SubSection{Registration with Adaptive Threshold}
Given a sequence of face images $\textbf{I}=\{I_{1} \cdots I_{t}\cdots I_{T}\}$, the identity labels $\textbf{P}=\{P_{1} \cdots P_{t} \cdots P_{T}\}$ and the embeddings $\textbf{F}=\{F_{1} \cdots F_{t} \cdots F_{T}\}$ extracted by the deep CNN, we register the embeddings into the database one-by-one as illustrated in Figure~\ref{fig:register}. At each registration $t$, we insert the feature embedding $F_{t}$ and its identity $P_{t}$ into the database; then we update the threshold $\sigma_{t}^{\tau}$ accordingly. $\sigma_{t}^{\tau}$ denotes the threshold of registered feature embedding $F_{\tau}$ when $F_{t}$ is registering into the database ($\tau = 1, \cdots, t$).

We assign a different threshold for each facial embedding in the database. For threshold $\sigma_{t}^{\tau}$ (associated with the $\tau$-th image at the $t$-th time), we first compute the similarity score between the embeddings $F_{\tau}$ and $F_{v}$ in the database ($v=1\cdots t$):
\begin{equation}
    S(\tau, v) = F_{\tau} \cdot F_{v}
\end{equation}
Then we compute $\sigma_t^\tau$ as the maximum value among all facial embeddings not belonging to the same person at the current time:
\begin{equation}
\begin{aligned}
    \sigma_{t}^{\tau} = \max(S(\tau, v))
    ,~v=1 \cdots t
    ,~\mathrm{where}~P_{\tau} \neq P_{v}
\end{aligned}
\end{equation}
For the implementation of updating the thresholds, as the face images are registered one at a time, we can take advantage of the recursion of $\sigma_{t-1}^{\tau}$ and $\sigma_{t}^{\tau}$ for an efficient computation during registration.


\SubSection{Recognition and authentication}

\begin{figure}[t]
\centering
\includegraphics[scale=0.2,keepaspectratio]{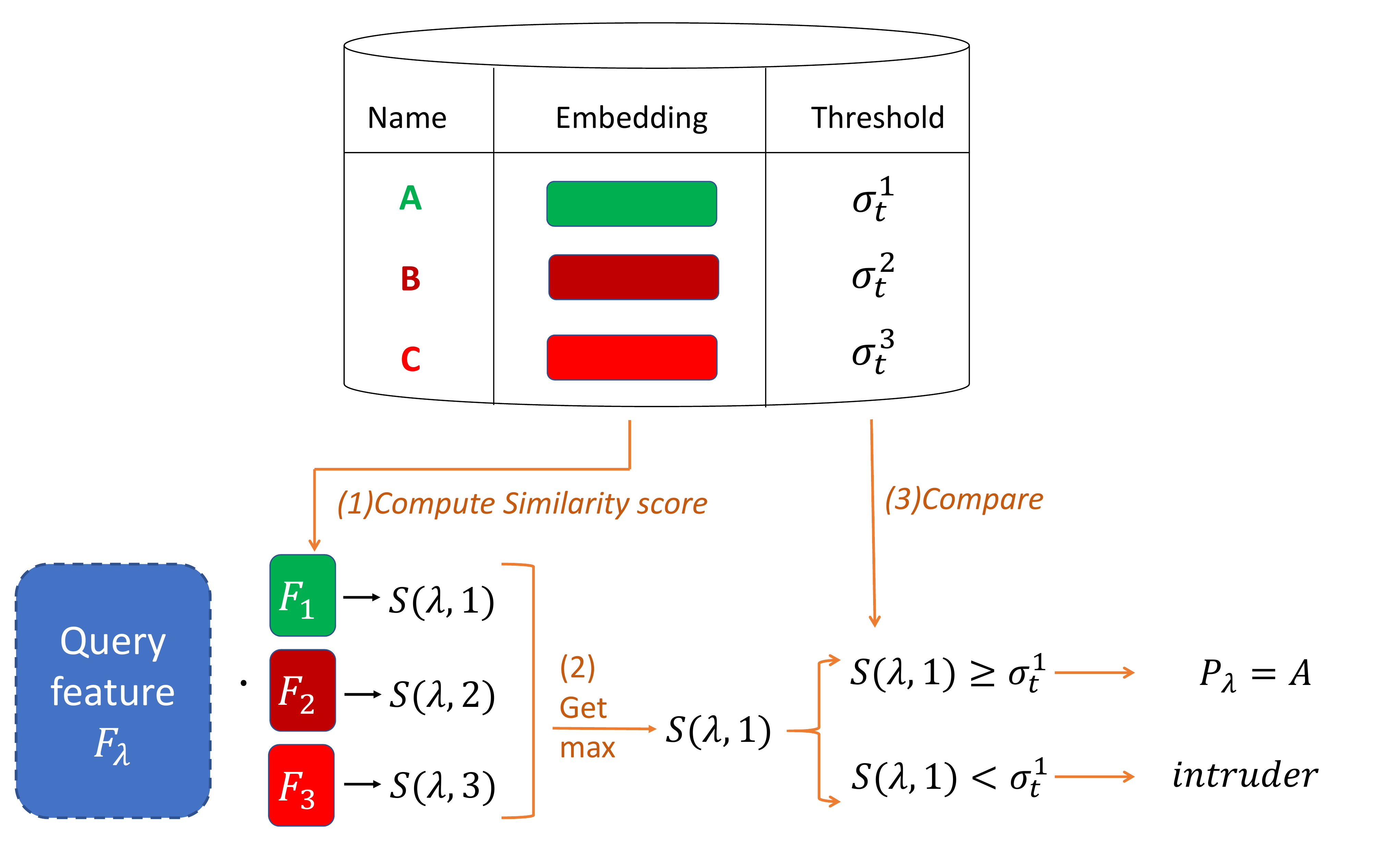}
\caption{Recognizing Flowchart. $F_{\lambda}$ is the embedding of a query image $I_{\lambda}$. (1) Compute the similarity score of $F_{\lambda}$ with all embeddings. (2) Get the maximal similarity score. (3) Compare the score with the stored threshold to determine whether the query is an intruder or a registered identity.}
\label{fig:recognition}
\end{figure}

Given a query facial image $I_{\lambda}$ without its identity label, we first extract the embedding $F_{\lambda}$ by the deep CNN. Then we compute the similarity scores with all the embeddings that are already stored in the database as shown in Figure~\ref{fig:recognition}. We extract the one which has the highest similarity score with $F_{\lambda}$ and denote its index as $u$:
\begin{equation}
u = \argmax_{v} (S(\lambda, v)),~
\mathrm{for}~v=1 \cdots t
\end{equation}

Once the most similar embedding $F_{u}$ is found, the system compare the associated threshold $\sigma_{t}^{u}$ with the similarity score $S(\lambda, u)$. If $S(\lambda, u) \geq \sigma_{t}^{u}$, then the image $I_{\lambda}$ will be classified as identity $P_{u}$; else it dissociates from any registered identities, in this case, we call it an \textit{intruder} and reject the authentication request:
\begin{equation}
    P_{\lambda}=
    \begin{cases}
    P_{u},~\mathrm{if} \: 
    S(\lambda, u) \geq \sigma_{t}^{u}
    \\
    {intruder},~\mathrm{if} \: 
    S(\lambda, u) < \sigma_{t}^{u}
    \end{cases}
\end{equation}

\vspace{-12pt}
\Section{Evaluation and Experiment}
Current open-set identification protocol such as \cite{Klare_2015_CVPR} evaluates the algorithm under determinate probe (or query) and gallery (or database) sets. This setting standardizes the evaluation and makes the result measurable, but it does not consider the intruders nor changes in the gallery, which often occurs in industrial applications. 

We introduce \textit{timeline} in our evaluation protocol. At $t$-th time, we present an image to the system and check the correctness; we move the face image from the probe to the gallery set one at a time to simulate the real registration rocess and the changes in the gallery.

\SubSection{Evaluation protocol for singly registering and recognizing}

In our evaluation protocol, we have $T$ images in the testing data, the gallery begins with an empty set. When a probe feature embedding $F_{t}$ is presented to the system, it will be used to calculate the similarity scores with all the gallery feature embeddings $F_{\tau}, \tau=1 \dots t$. The gallery feature embedding that gets the highest similarity score with $F_{t}$ is denoted as $F_{u}$. The threshold function is denoted as $\Phi(t,u)$. For fixed threshold, $\Phi(t,u)$ always returns a constant value; for adaptive threshold, $\Phi(t,u)=\sigma_{t}^{u}$.
We use 10-fold CV to compute the fixed threshold can compare the results with those that obtained using our adative thresholding method.

When $F_{t}$ is presented to the system, if $S(t,u) \geq \Phi(t,u)$, and $P_{t}=P_{u}$, then we define this case as \textit{true accept}:
\begin{equation}
\mathrm{TA}(t) = \{S(t,u) \geq \Phi(t,u), \mathrm{and} \: P_{t}=P_{u} \}
\end{equation}

If $P_{t}$ is an identity in the gallery but $S(t,u) < \Phi(t,u)$, then no matter whether $P_{t}=P_{u}$, we define these cases as \textit{false reject}:
\begin{equation}
\mathrm{FR}(t) = \{S(t,u) < \Phi(t,u), \mathrm{and} \: P_{t} \in \textbf{P} \}
\end{equation}

If $P_{t}$ is not contained in the gallery, then we call it an \textit{intruder}. Thus, we define \textit{false accept} and \textit{true reject} as:
\begin{equation}
\mathrm{FA}(t) = \{S(t,u) \geq \Phi(t,u), \mathrm{and} \: P_{t} \notin \textbf{P} \}
\end{equation}
\begin{equation}
\mathrm{TR}(t) = \{S(t,u) < \Phi(t,u), \mathrm{and} \: P_{t} \notin \textbf{P} \}
\end{equation}

For the last possible case, if $S(t,u) \geq \Phi(t,u)$ and $P_{t}$ is someone in the gallery, but $P_{t}$ and $P_{u}$ are different identities, then we define this case as \textit{identification error}:
\begin{equation}
\begin{aligned}
    \mathrm{IE}(t) = \{S(t,u) \geq \Phi(t,u)
    &,~\mathrm{where}~P_{t} \neq P_{u} \\ &,~\mathrm{and}~P_{t} \in \textbf{P} \}
\end{aligned}
\end{equation}

We add $F_{t}$ to the gallery set iteratively to simulate the registration operation ($t=1\cdots T$); we extract $F_{t+1}$ from the probe set and repeat the process until all of the $T$ images are examined. The final accuracy, $\mathrm{ACC}$, 
is then defined as the average correctness of all the $T$ sessions:
\begin{equation}
\label{ACC}
\mathrm{ACC}=\frac{\sum_{t=1}^T |\mathrm{TA}(t)|+|\mathrm{TR}(t)|}{T}
\end{equation}

\begin{table}[ht]
\centering
\vspace{-10pt}
\caption{Summary of the evaluation protocol}
\label{tab:evaluation-protocol}
\begin{tabular}{lll}
\toprule
& $S(t,u) \geq \Phi(t,u)$
& $S(t,u) < \Phi(t,u)$   \\ \midrule
$P_{t}=P_{u}$        & True accept &  False reject \\ \midrule
$P_{t} \neq P_{u},$  & \multirow{2}{8em}{Identification error}  &  \multirow{2}{5em}{False reject} \\
$P_{t} \in \textbf{P}$ &  &  \\ \midrule
$P_{t} \neq P_{u},$  & \multirow{2}{8em}{False accept }  &  \multirow{2}{5em}{True reject} \\
$P_{t} \notin \textbf{P}$ &  &  \\ \midrule
\end{tabular}
\end{table}
\vspace{-10pt}

\SubSection{Experiments on Facial Datasets}
In all our experiments we use the same deep CNN trained on MS-Celeb-1M~\cite{guo2016msceleb} dataset to extract the feature embedding. We evaluate our algorithm under the aforementioned protocol on three datasets: Labeled Faces in the Wild (LFW)~\cite{huang2008labeled}, Adience~\cite{eidinger2014age} and Color FERET~\cite{phillips2000feret}. As some faces in the images are undetectable, we did not use all images in the datasets. The statistics of the images used in our experiments are show in Table~\ref{tab:dataset-info}. For each experiment, we randomly shuffle all of the images in the dataset, and then register the images one-by-one according to the random order. We perform the experiments 10 times on each dataset and average the accuracy. To fairly compare with the results of using a fixed threshold, we also conduct the fixed-thresholding experiments under our evaluation protocol with the same registration order.

\begin{table}[t]
\centering
\caption{The number of aligned images, identities, and number of samples per identity (with the standard deviation) in facial datasets.} 
\label{tab:dataset-info}
\begin{tabular}{lrrr}
\toprule
           &\#images & \#classes & \#images/class\\ \midrule
LFW         & 13,233 & 5,749 & $2.3\pm9.01$\\ 
Adience     & 19,339 & 2,284 & $8.46\pm23.13$\\ 
Color FERET & 11,285 & 994   & $11.35\pm8.6$\\ \midrule
\end{tabular}
\end{table}

As for the selection of fixed thresholds, following the verification benchmark of LFW, we use 6,000 image pairs randomly generated from the dataset. As done in the protocol of LFW for the verification performance evaluation, 10-fold CV is used and a threshold is selected for each fold-splitting. The 10 thresholds selected are then averaged to yield the fixed threshold used in our experiment. The same procedure is performed for the fixed thresholds selection of the color FERET and Adience datasets too.
After that, the fixed thresholds also serve as the initial values ($t=1$) of our adaptive thresholding procedure.



\begin{figure}[t]
\centering
\vspace{-10pt}
\includegraphics[trim={2.5cm 0 3cm 0},clip,height=4cm]{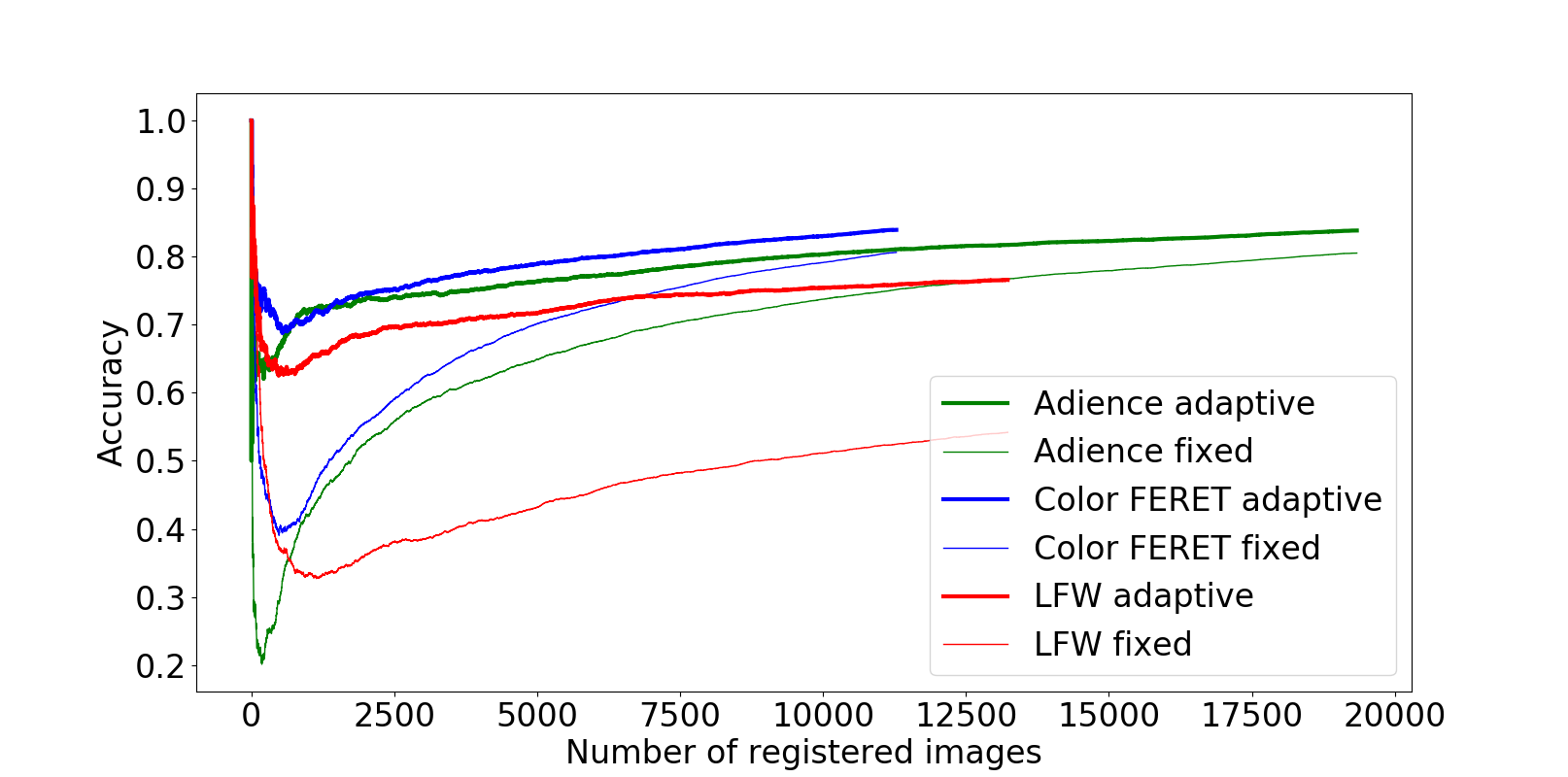}
\caption{Comparisons of temporary accuracy during each registration.}
\label{fig:result}
\end{figure}

The accuracy (ACC defined in Eq.~\ref{ACC}) is shown in Table~\ref{tab:experiment-result}.
In the case of fewer samples per identity like LFW, the adaptive-threshold approach outperforms the fixed-threshold method by around 22.5\% in accuracy. In Adience or Color FERET, the adaptive-threshold approach still consistently outperforms the fixed one.
The temporary accuracy during each registration is shown in Figure~\ref{fig:result}. The adaptive-threshold approach converges earlier, showing it's promising in sample limited applications, e.g., small factory security systems.

\Section{Conclusion}
We solve the long-term threshold-selection problem for face recognition and authentication, and also tackle the deficiency of current evaluation protocol. The introduced technique of adaptive thresholding can decide more favorable thresholds. We also design an on-line registration system for real-world scenarios, which can handle varied conditions for practical applications.

\begin{table}[t]
\centering
\caption{Final Accuracy}
\label{tab:experiment-result}
\begin{tabular}{lcc}
\toprule
            &  Adaptive & Fixed / Threshold \\ \midrule
LFW         & $\textbf{76.46}\% $ & $53.97\%/0.3779$\\
Adience     & $\textbf{84.30}\% $ & $80.60\% /0.2487 $\\
Color FERET & $\textbf{83.79}\% $ & $80.72\%/0.3968$\\ \midrule
\end{tabular}
\end{table}


\nocite{}
\bibliographystyle{latex8}
\bibliography{latex8}

\begin{thebibliography}{10}\setlength{\itemsep}{-1ex}\small

\bibitem{chen2018mobilefacenets}
S.~Chen, Y.~Liu, X.~Gao, and Z.~Han.
\newblock Mobilefacenets: Efficient cnns for accurate real-time face
  verification on mobile devices.
\newblock In {\em CCBR}, 2018.

\bibitem{eidinger2014age}
E.~Eidinger, R.~Enbar, and T.~Hassner.
\newblock Age and gender estimation of unfiltered faces.
\newblock {\em IEEE TIFS}, 2014.

\bibitem{guo2016msceleb}
Y.~Guo, L.~Zhang, Y.~Hu, X.~He, and J.~Gao.
\newblock M{S}-{C}eleb-1{M}: A dataset and benchmark for large scale face
  recognition.
\newblock In {\em ECCV}, 2016.

\bibitem{huang2008labeled}
G.~B. Huang, M.~Mattar, T.~Berg, and E.~Learned-Miller.
\newblock Labeled faces in the wild: A database forstudying face recognition in
  unconstrained environments.
\newblock In {\em ECCV Workshop}, 2008.

\bibitem{Klare_2015_CVPR}
B.~F. Klare, B.~Klein, E.~Taborsky, A.~Blanton, J.~Cheney, K.~Allen,
  P.~Grother, A.~Mah, and A.~K. Jain.
\newblock Pushing the frontiers of unconstrained face detection and
  recognition: Iarpa janus benchmark a.
\newblock In {\em CVPR}, 2015.

\bibitem{liu2017sphereface}
W.~Liu, Y.~Wen, Z.~Yu, M.~Li, B.~Raj, and L.~Song.
\newblock Sphereface: Deep hypersphere embedding for face recognition.
\newblock In {\em IEEE CVPR}, 2017.

\bibitem{parkhi2015deep}
O.~M. Parkhi, A.~Vedaldi, A.~Zisserman, et~al.
\newblock Deep face recognition.
\newblock In {\em BMVC}, 2015.

\bibitem{phillips2000feret}
P.~J. Phillips, H.~Moon, S.~A. Rizvi, and P.~J. Rauss.
\newblock The feret evaluation methodology for face-recognition algorithms.
\newblock {\em IEEE TPAMI}, 2000.

\bibitem{schroff2015facenet}
F.~Schroff, D.~Kalenichenko, and J.~Philbin.
\newblock Facenet: A unified embedding for face recognition and clustering.
\newblock In {\em CVPR}, 2015.

\bibitem{simonyan2014very}
K.~Simonyan and A.~Zisserman.
\newblock Very deep convolutional networks for large-scale image recognition.
\newblock {\em ICLR}, 2015.

\bibitem{szegedy2017inception}
C.~Szegedy, S.~Ioffe, V.~Vanhoucke, and A.~A. Alemi.
\newblock Inception-v4, inception-resnet and the impact of residual connections
  on learning.
\newblock In {\em AAAI}, 2017.

\bibitem{wang2018additive}
F.~Wang, J.~Cheng, W.~Liu, and H.~Liu.
\newblock Additive margin softmax for face verification.
\newblock {\em IEEE SPL}, 2018.

\bibitem{wang2017normface}
F.~Wang, X.~Xiang, J.~Cheng, and A.~L. Yuille.
\newblock Normface: l 2 hypersphere embedding for face verification.
\newblock In {\em ACMMM}, 2017.

\bibitem{zhang2016joint}
K.~Zhang, Z.~Zhang, Z.~Li, and Y.~Qiao.
\newblock Joint face detection and alignment using multitask cascaded
  convolutional networks.
\newblock {\em IEEE SPL}, 2016.

\end{thebibliography}

\end{document}